\documentclass{article}
\usepackage[T1]{fontenc}
\usepackage{iclr2027_conference,times}

\usepackage{amsmath,amsfonts,bm}

\def\eqref#1{equation~\ref{#1}}
\def\1{\bm{1}}

\DeclareMathAlphabet{\mathsfit}{\encodingdefault}{\sfdefault}{m}{sl}
\SetMathAlphabet{\mathsfit}{bold}{\encodingdefault}{\sfdefault}{bx}{n}

\usepackage{amsmath}
\usepackage{amssymb}
\usepackage{booktabs}
\usepackage{float}
\usepackage{graphicx}
\usepackage{placeins}
\usepackage{microtype}
\usepackage{xcolor}
\usepackage{tikz}
\usepackage{xspace}
\usepackage{hyperref}
\usepackage{url}
\usetikzlibrary{arrows.meta,backgrounds,calc,fit,positioning}
\hypersetup{hidelinks}

\definecolor{BackboneBlue}{HTML}{DCEBFA}
\definecolor{MemoryGold}{HTML}{F9E7B4}
\definecolor{HeadGreen}{HTML}{DDF1E3}
\definecolor{ControlRose}{HTML}{F5DADF}
\definecolor{Ink}{HTML}{263238}

\newcommand{\retrievetoken}{\texttt{<retrieve>}\xspace}

\title{PERSIST: Who-What-When Memory Across Sessions for Full-Duplex Spoken Dialogue}

\author{%
{\small Achira Lin$^{1,*}$\quad Siyuan Hou$^{1,*}$\quad Wenyi Yu$^{1}$\quad Xinnian Zhao$^{1}$\quad Haoyu Niu$^{1}$}\\[0.45em]
{\small Gengwang$^{2}$\quad Longshuai Xiao$^{2}$\quad Shihai Xiao$^{2}$\quad Mangsuo Zhao$^{1}$\quad Chao Zhang$^{1,\dagger}$}\\[0.8em]
{\small $^{1}$Tsinghua University\qquad $^{2}$Huawei Technologies Ltd.}\\[0.35em]
{\footnotesize $^{*}$Equal contribution.\quad $^{\dagger}$Corresponding author: \texttt{cz277@tsinghua.edu.cn}}
}

\makeatletter
\renewcommand{\@maketitle}{%
  \begin{center}
    {\LARGE\bfseries \@title\par}
    \vskip 1.2em
    {\normalsize\normalfont \@author\par}
  \end{center}
  \vskip 0.6em
}
\makeatother
\renewcommand{\headrulewidth}{0pt}
\iclrfinalcopy % Show authors and disable review line numbers.
\usepackage{wrapfig}
\usepackage{capt-of}
\begin{document}

\raggedbottom

\maketitle

\begin{abstract}

%Modern voice assistants may be shared by multiple users and should be able to answer questions about earlier conversations, such as ``When did I originally plan to leave?'', or adapt their behavior to individual users based on past interactions. 
Modern voice assistants may be shared by multiple users and should be able to answer questions about earlier conversations such as ``When did I originally plan to leave?'' or adapt their behavior to individual users based on past interactions.
This requires more than retrieving a topically similar passage: the assistant must identify the current speaker, recover the relevant past state, and distinguish it from later revisions. We present PERSIST, a persistent memory system for multi-session, multi-speaker spoken dialogue that explicitly models \textit{Who}, \textit{What}, and \textit{When}. PERSIST structures cross-session histories into readable event records and retrieves them with a 3W joint scoring mechanism that combines semantic content, acoustic speaker identity, and temporal state. For real-time full-duplex interaction, PERSIST further reuses intermediate representations from the dialogue backbone, avoiding query-audio re-encoding and reducing retrieval latency from 578.42 ms to 7.03 ms. We also introduce SpokenTrace, a diagnostic benchmark that factorizes evaluation along memory tasks and speaker-query types, exposing failures in recall, speaker attribution, and temporal-state tracking. On SpokenTrace, PERSIST achieves 85.08\% end-to-end task accuracy and improves all-support EM@3 from 49.01\% with BGE-large to 82.10\%.

\end{abstract}
\section{Introduction}
\label{sec:introduction}

%Persistent long-term memory is fundamental for conversational agents to maintain continuity, track user preferences, and preserve evolving context across multi-session interactions. 
Persistent long-term memory is fundamental for conversational agents to maintain continuity, adapt to individual users, and preserve evolving context across multi-session interactions.
While text-based long-term memory has been widely explored~\citep{packer2024memgptllmsoperatingsystems, xu2025amemagenticmemoryllm, chhikara2025mem0buildingproductionreadyai}, extending persistent memory to spoken dialogue requires grounding these evolving conversational states in acoustic speaker identities across sessions. In scenarios such as collaborative meetings, family assistance, or group planning, plans and commitments continually evolve as different participants contribute updates, revisions, or counter-proposals over time. Consequently, an effective spoken memory system must retrieve not only \textit{what} was said, but also \textit{who} said it and \textit{when} the relevant state was active.
%Consequently, an effective spoken memory system must retrieve not only \textit{what} was said (semantic content), but also \textit{who} said it (speaker attribution) and \textit{when} the relevant state was active (temporal state tracking).

%an effective spoken memory system cannot rely solely on semantic content matching; it must simultaneously perform \textbf{speaker attribution} (\textit{who} stated or modified a fact) and track \textbf{temporal state evolution} (\textit{when} a specific state was active).

\begin{figure*}[t] 
\centering
\includegraphics[width=\textwidth,] %trim={7mm 35mm 75mm 12mm},clip]
{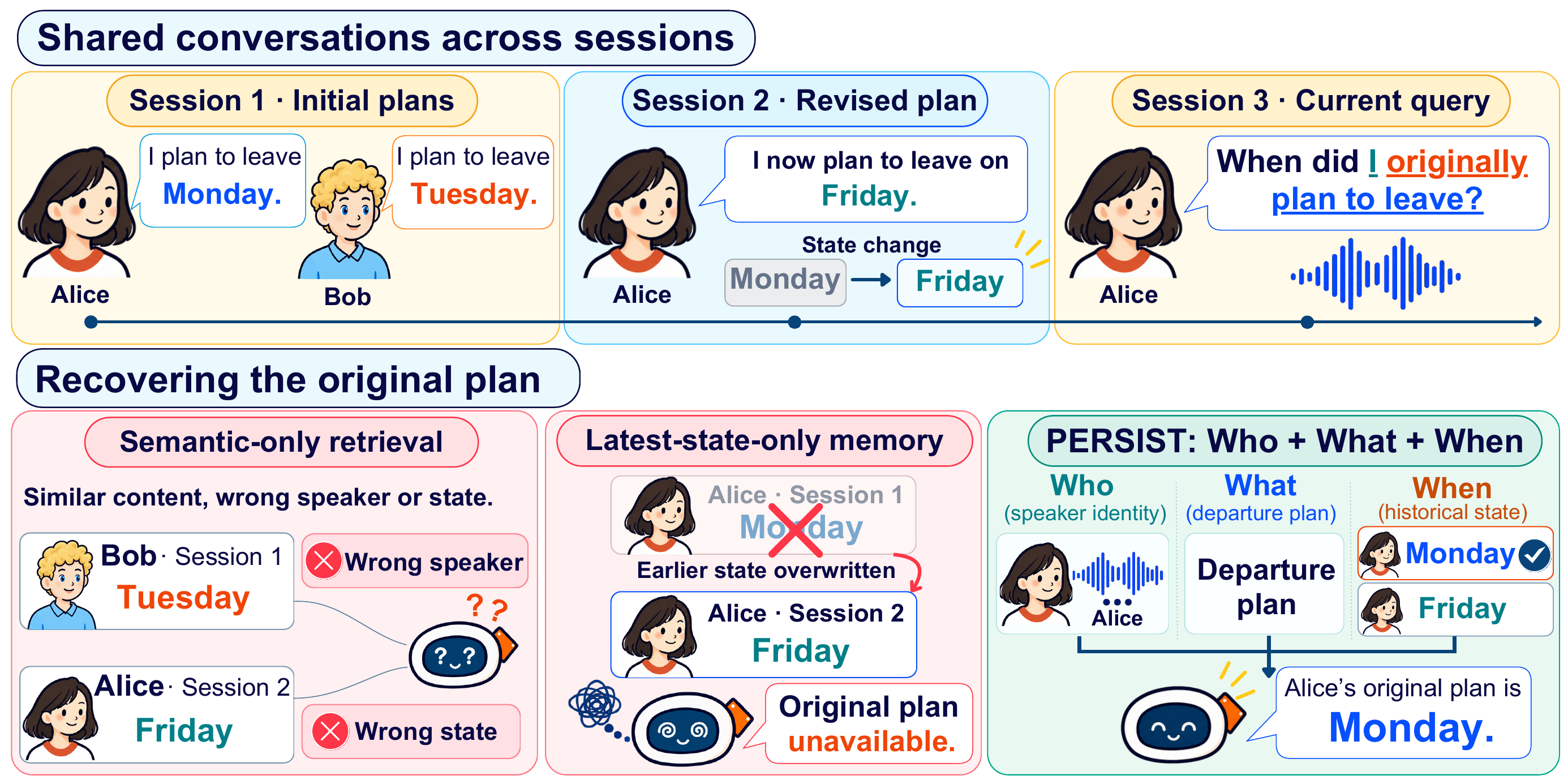}
\caption{\textbf{Motivation and failure modes of existing speech memory paradigms under dynamic multi-speaker dialogue.}}
\label{fig:motivation}
\end{figure*}

%To address these requirements, existing speech memory systems often combine semantic retrieval with structured fact updates, semantic retrieval and structured fact updates, both of which encounter key limitations in dynamic multi-speaker scenarios (Figure~\ref{fig:motivation}). The first approach, centered on text-based semantic search \citep{Zulfikar_2024}, struggles with speaker and temporal ambiguities when query content overlaps across participants or sessions. Fact-updating mechanisms that prioritize the latest user state~\citep{xie2026voicememstreamingdualbrainmemory}, can overwrite or obscure earlier states needed for retrospective queries. Crucially, when speaker and temporal metadata are used solely as hard filtering constraints, they determine which records are eligible for retrieval without directly contributing to graded relevance scores. This motivates incorporating acoustic speaker signals and query-dependent temporal intent into relevance scoring alongside semantic content.

Existing spoken memory systems commonly rely on semantic retrieval or structured fact updates, both of which face limitations in dynamic multi-speaker scenarios (Figure~\ref{fig:motivation}). Semantic retrieval \citep{Zulfikar_2024} can recover topically relevant content but may fail to distinguish between different speakers or temporal states when similar information recurs across participants or sessions. 
Structured fact-updating mechanisms \citep{xie2026voicememstreamingdualbrainmemory}, by contrast, often prioritize the latest state and may overwrite earlier states needed for retrospective queries. These limitations motivate memory retrieval that jointly considers \textit{Who}, \textit{What}, and \textit{When}, rather than treating speaker and time merely as filtering metadata.

%To bridge this gap, we present \textsc{Persist}, a persistent memory system that unifies Who (speaker identity), What (semantic content), and When (temporal state) into a single representation space for multi-speaker spoken dialogue. \textsc{Persist} structures cross-session histories into explicit, human-readable event records grounded in speaker identities and utterance timestamps. During query--memory matching, Its 3W scoring mechanism combines speaker-conditioned semantic matching with temporal relevance in a single retrieval operation.

To bridge this gap, we present \textsc{Persist}, a persistent memory system for multi-session, multi-speaker spoken dialogue that explicitly models \textit{Who} (speaker identity), \textit{What} (semantic content), and \textit{When} (temporal state). \textsc{Persist} structures cross-session histories into readable event records and retrieves them with a 3W joint scoring mechanism that combines semantic content, acoustic speaker identity, and temporal state. 
This memory problem becomes particularly demanding in full-duplex spoken dialogue~\citep{defossez2024moshi}. Continuous, simultaneous listening and speaking rapidly exhaust the active context, making persistent memory increasingly important, while natural conversational flow places a tight latency budget on memory retrieval. To meet these dual demands, we instantiate \textsc{Persist}'s backbone-native path within SALMONN-omni~\citep{yu2025salmonnomni}. By reusing intermediate acoustic features and hidden states already computed during forward generation, \textsc{Persist} avoids re-encoding query audio at retrieval time, reducing memory-retrieval latency to 7.03~ms.

%Beyond retrieval accuracy, real-world deployment requires memory systems to be both architecturally flexible and capable of operating under strict real-time constraints. While \textsc{Persist} offers a modular external path adaptable to standard dialogue pipelines, next-generation full-duplex speech models~\citep{defossez2024moshi, yu2025salmonnomni} present an exceptionally demanding testbed. In full-duplex interaction, continuous simultaneous listening and speaking rapidly exhaust the active context, making persistent memory indispensable, while natural conversational flow places a tight latency budget on memory retrieval. To meet these dual demands, we instantiate \textsc{Persist}'s backbone-native path within SALMONN-omni~\citep{yu2025salmonnomni}. By reusing intermediate acoustic features and hidden states already computed during forward generation, \textsc{Persist} avoids re-encoding query audio at retrieval time, reducing memory-retrieval latency to an ultra-low 7.03~ms.

Evaluating persistent, speaker-attributed memory requires benchmarks that jointly assess multi-session history, speaker identity, and temporal state evolution. While existing benchmarks explore long-term recall in text~\citep{maharana-etal-2024-evaluating, wu2025longmemevalbenchmarkingchatassistants}, multi-party information updating~\citep{hu2026evaluatinglonghorizonmemorymultiparty, yang2026groupmembenchbenchmarkingllmagent}, or spoken dialogue memory~\citep{sun2026msubenchspeakercentricunderstandingconversational, xie2026voicememstreamingdualbrainmemory}, none systematically isolate how cross-session acoustic speaker grounding interacts with temporal-state retrieval. To fill this gap, we introduce \textsc{SpokenTrace}, which factorizes evaluation along two complementary axes---memory tasks and speaker-query types---to separately and jointly diagnose memory retrieval, speaker attribution, and temporal tracking. On SpokenTrace, PERSIST achieves an end-to-end task accuracy of 85.08\% and improves all-support EM@3 from 49.01\% with BGE-large to 82.10\%.

%a multi-session spoken-dialogue benchmark that organizes queries along two complementary axes: 

Our primary contributions are summarized as follows:
\begin{itemize}
    \item \textbf{Problem Characterization:} We characterize the joint requirements of semantic-content retrieval, speaker attribution, and temporal-state retrieval in multi-session spoken dialogue.
    %We characterize the joint requirements of speaker attribution and temporal-state retrieval in multi-session spoken memory.
    \item \textbf{The \textsc{Persist} Framework:} We introduce \textsc{Persist}, which writes readable, speaker- and time-grounded event records and retrieves them with 3W joint scoring over \textit{Who}, \textit{What}, and \textit{When}. For real-time full-duplex interaction, \textsc{Persist} reuses dialogue-backbone representations to avoid redundant query-audio encoding, reducing retrieval latency to $7.03\text{~ms}$.

    %\item \textbf{The \textsc{Persist} Framework:} We introduce \textsc{Persist}, which writes readable, speaker- and time-grounded event records and retrieves them with 3W joint scoring over \textit{Who}, \textit{What}, and \textit{When}. For real-time full-duplex interaction, \textsc{Persist} further reuses dialogue-backbone representations to avoid redundant query-audio encoding, reducing retrieval latency to $7.03\text{~ms}$.
    %a persistent memory architecture featuring a 3W joint scoring mechanism that unifies semantic content, acoustic speaker embeddings, and temporal representations. Furthermore, we demonstrate that directly reusing intermediate backbone representations eliminates redundant encoding, enabling ultra-low retrieval latency ($7.03\text{~ms}$) for full-duplex systems.
    %\item \textbf{The \textsc{Persist} Framework:} We introduce \textsc{Persist}, a persistent memory architecture featuring a 3W joint scoring mechanism that unifies semantic content, acoustic speaker embeddings, and temporal representations. Furthermore, we demonstrate that directly reusing intermediate backbone representations eliminates redundant encoding, enabling ultra-low retrieval latency ($7.03\text{~ms}$) for full-duplex systems.
    \item \textbf{The \textsc{SpokenTrace} Benchmark:} We design and release \textsc{SpokenTrace}, a multi-session diagnostic benchmark that orthogonally evaluates speaker attribution, temporal state evolution, and memory recall, providing a standardized evaluation testbed for long-term spoken dialogue agents.
\end{itemize}
\section{PERSIST: A Who-What-When Memory Framework}
\label{sec:system}

\subsection{Framework Overview}
\label{sec:framework_overview}

A spoken-dialogue assistant equipped with \textsc{PERSIST} uses two complementary forms of memory: a short-term active context for ongoing interaction and a long-term persistent store for conversational history. The active context gives the dialogue model direct access to recent turns, while the persistent store preserves readable conversational records beyond the current context and across sessions.
% We focus on the persistence, attribution, and retrieval of conversational records; selective memory formation, consolidation, and forgetting are outside the scope of this work.
At a high level, PERSIST performs two core operations: writing speaker- and time-grounded conversational events into persistent memory, and retrieving them through Who-What-When matching.

\begin{figure}[!htbp]
    \centering
    \includegraphics[width=1\linewidth,]%trim={7mm 8mm 3mm 25mm},clip]
        {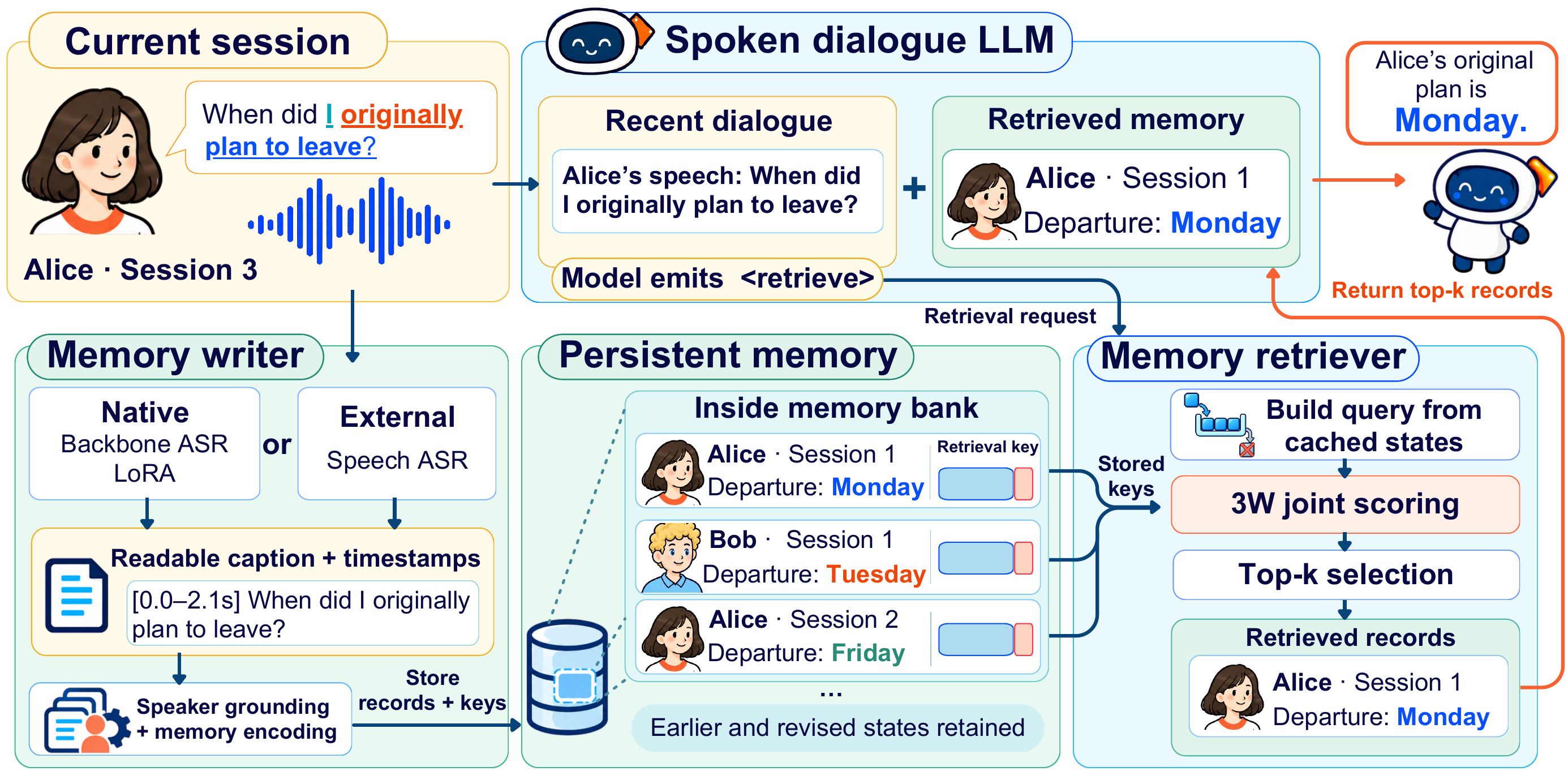}
    \caption{\textbf{PERSIST system overview.}
The memory writer continuously converts speech spans into readable memory records that persist beyond the active context and across sessions. When the dialogue model emits the special token \texttt{\textless{}retrieve\textgreater{}}, the retriever encodes a query and scores it against stored keys using semantic, speaker, and temporal information. The top-$k$ records are then incorporated into the model's short-term active context for response generation.}
    \label{fig:salmonn-omni-memory}
\end{figure}

The memory interface is independent of how records are produced. In this work, we instantiate \textsc{PERSIST} with two writing paths: a backbone-native path that reuses intermediate representations from the dialogue model, and a plug-and-play external path based on a speaker-attributed speech recognition system. Both paths produce the same readable event-record format and share the same speaker grounding, persistent storage, and 3W retrieval pipeline.

%The memory interface is independent of how records are produced or retrieved. In this work, we instantiate \textsc{PERSIST} with two complementary writing paths: a \emph{backbone-native} path that reuses intermediate representations from the dialogue model, and a \emph{plug-and-play external} path based on a speaker-attributed speech recognition system. Both paths produce the same readable record format, preserving content, speaker identity, time, and session provenance, while sharing the same persistent storage and downstream retrieval pipeline.

%\subsection{Persistent Retrieval and Full-Duplex Memory Access}
\subsection{Who-What-When Retrieval and Full-Duplex Memory Access}
\label{sec:memory_retrieval} 
%Full-duplex models can generate responses while receiving the user's speech. With input and output proceeding in parallel, the model must process incoming speech while generating responses and handling possible interruptions or corrections, so the interaction history accumulates continuously and fills the limited active context more quickly than in turn-based interaction. Memory retrieval must also be fast enough to provide historical information during the conversation without causing delays or making the interaction feel unnatural. At the same time, full-duplex systems can operate in a half-duplex or turn-based mode when speakers do not overlap, whereas systems designed only for turn-based interaction may not support simultaneous listening and speaking. For these reasons, we implement PERSIST in SALMONN-omni~\cite{yu2025salmonnomni} and use full-duplex speech interaction as a more demanding validation setting that covers a broader range of interaction patterns, while keeping PERSIST's memory interface independent of the underlying model's interaction mode.

\begin{figure}[!htbp]
    \centering
    \includegraphics[width=\linewidth,]%trim={9mm 12.5mm 9mm 32.5mm},clip]
    {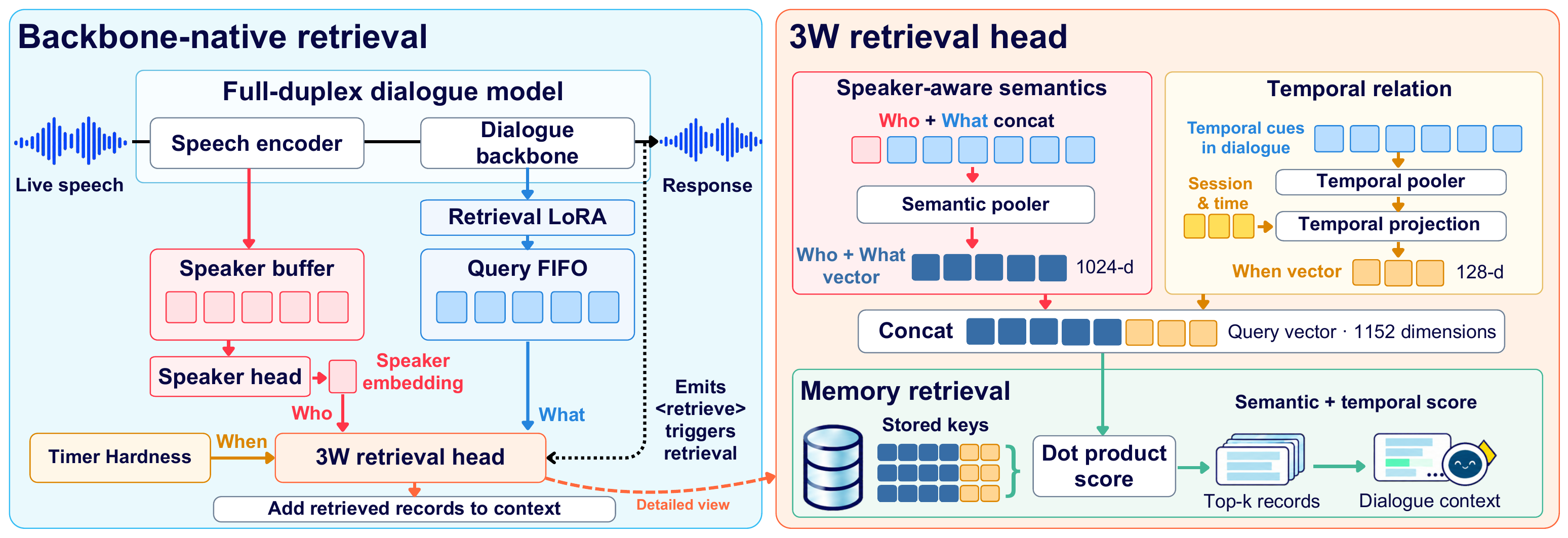}
    \caption{\textbf{Backbone-native implementation of PERSIST.}
During memory writing (lower route), an ASR-specific LoRA first generates a readable caption for each completed speech span, which the backbone then encodes into a semantic representation using retrieval-specific LoRA. A speaker head extracts speaker embeddings from intermediate speech-encoder features cached in the Speaker Buffer. Finally, the Semantic--Temporal Pooler fuses the speaker-aware semantic representation with temporal features to form a 1,152-dimensional memory vector, which is stored alongside its caption in the CPU memory store. During retrieval (upper route), recent speech representations are incrementally cached in a Query FIFO (first-in, first-out) buffer. When the dialogue model emits the retrieval trigger, the Semantic-Temporal Pooler combines the cached representations, the speaker embedding extracted from the Speaker Buffer, and temporal information to form a 1,152-dimensional query vector. Inner-product top-$k$ search ranks the stored memory vectors and returns their corresponding captions to the dialogue context.}
    \label{fig:backbone-native-detail}
\end{figure}

Full-duplex models must process incoming speech while generating responses and handling interruptions or corrections, causing interaction history to accumulate continuously and consume the limited active context. At the same time, memory retrieval must remain fast enough not to disrupt real-time interaction. We therefore instantiate \textsc{PERSIST} within SALMONN-omni, using full-duplex speech interaction as a demanding real-time setting while keeping the memory interface independent of the underlying interaction mode.
Figure~\ref{fig:backbone-native-detail} illustrates how the backbone-native implementation writes speaker- and time-aware memory representations and retrieves relevant records from them.

\paragraph{Retrieval trigger.}
We make retrieval a native action of the dialogue model, just as a full-duplex model determines when to begin speaking. This avoids adding a separate decision model and its inference step to the response path, which could increase latency. Following the model-emitted retrieval-token approach of Self-RAG~\citep{asai2024selfrag}, we train the dialogue model to emit \retrievetoken when answering requires information beyond the active context. The token triggers memory retrieval, and the retrieved captions are added to the context before generation continues.

\paragraph{Retrieval encoding.}
To meet the full-duplex latency constraint described above, we use a
retrieval-specific LoRA on the shared dialogue backbone to adapt its
representations for query--memory matching. Because the retrieval
route computes query states incrementally, we cache its most recent
128 hidden states from decoder layer 19 in a FIFO buffer. When
retrieval is triggered, the retrieval head constructs the query from
these cached states without reprocessing the dialogue history.

Let $q$ denote the current query and $i$ index a candidate memory. We
use $\mathbf{H}_x$ for the hidden-state sequence associated with
$x\in\{q,i\}$: $\mathbf{H}_q$ contains the cached query states, while
$\mathbf{H}_i$ contains the encoded states of memory caption $i$.
Each sequence has the form
$\mathbf{H}_x\in\mathbb{R}^{N_x\times4096}$, where $N_x$ is its
number of states.

To let retrieval jointly account for semantic content, speaker
identity, and temporal information, we represent each query and
memory as a 1,152-dimensional vector comprising a
1,024-dimensional semantic component and a 128-dimensional temporal
component.

In our formulation, Who is incorporated into speaker-conditioned semantic matching, while When contributes an explicit temporal component.
For the semantic component, let $\mathbf{e}^{\mathrm{spk}}_x$ denote
the speaker embedding for $x\in\{q,i\}$. We prepend it to the
corresponding content states as an additional token, allowing the
pooled representation to encode both what was said and who said it. The concatenated state matrix has shape $\widetilde{\mathbf{H}}_x\in\mathbb{R}^{(N_x+1)\times4096}$.
We use a shared semantic attention pooler $P_{\mathrm{sem}}$ to map
the resulting states to 1024-dimensional vectors, which we L2-normalize
so that their inner product reflects cosine similarity rather than
vector magnitude:
\begin{equation}
\begin{alignedat}{2}
&\widetilde{\mathbf{H}}_x
&\quad &= \operatorname{Concat}_{\mathrm{tok}}
(\mathbf{e}^{\mathrm{spk}}_x,\mathbf{H}_x),
\qquad x\in\{q,i\},\\
&\mathbf{q}_{\mathrm{sem}}
&\quad &= \operatorname{norm}\!(
P_{\mathrm{sem}}(\widetilde{\mathbf{H}}_q)
),\\
&\mathbf{k}_{i,\mathrm{sem}}
&\quad &= \operatorname{norm}\!(
P_{\mathrm{sem}}(\widetilde{\mathbf{H}}_i)
),
\end{alignedat}
\end{equation}
where $\operatorname{norm}(\mathbf{v})=
\mathbf{v}/\|\mathbf{v}\|_2$.

For the temporal component, we encode time separately from semantic
content, following temporal conditioning in
TempRetriever~\citep{tempretriever} and temporal subspaces in
TMRL~\citep{tmrl}. The query and memory encodings are asymmetric:
a memory's temporal key represents when the event occurred, while the
query's content indicates the temporal relation being requested.
Accordingly, we derive each memory's temporal key directly from its session and
time features. On the query side, a temporal pooler aggregates the
states in $\mathbf{H}_q$ to capture the requested temporal relation.

Let $s_x$ and $t_x$ denote the session index and session-relative
time of query or memory $x$, respectively. We map them to fixed
six-dimensional features $\mathbf{f}(s_x,t_x)$, with $[\cdot;\cdot]$
denoting feature concatenation. Here, $P_{\mathrm{time}}$ is a
temporal attention pooler, and $W_q^{\mathrm{time}}$ and
$W_k^{\mathrm{time}}$ denote learned query and memory mappings,
respectively. We apply
$\phi(\mathbf{x})=\mathbf{x}/(1+\|\mathbf{x}\|_2)$
to bound the temporal representations before scoring:
\begin{equation}
\begin{alignedat}{2}
&\mathbf{q}_{\mathrm{time}}
&\quad &= \phi\!\left(
W_q^{\mathrm{time}}
[P_{\mathrm{time}}(\mathbf{H}_q);
\mathbf{f}(s_q,t_q)]
\right),\\
&\mathbf{k}_{i,\mathrm{time}}
&\quad &= \phi\!\left(
W_k^{\mathrm{time}}\mathbf{f}(s_i,t_i)
\right).
\end{alignedat}
\end{equation}

Finally, we concatenate the semantic and temporal components without
further normalization:
\begin{equation}
\mathbf{q} = [\mathbf{q}_{\mathrm{sem}};\mathbf{q}_{\mathrm{time}}],
\qquad
\mathbf{k}_i = [\mathbf{k}_{i,\mathrm{sem}};\mathbf{k}_{i,\mathrm{time}}].
\end{equation}
The retrieval score is therefore
\begin{equation}
r_i = \mathbf{q}^{\top}\mathbf{k}_i
= \mathbf{q}_{\mathrm{sem}}^{\top}\mathbf{k}_{i,\mathrm{sem}}
+ \mathbf{q}_{\mathrm{time}}^{\top}\mathbf{k}_{i,\mathrm{time}}.
\end{equation}

%\section{SpokenTrace: A Diagnostic Benchmark for Multi-Session Speaker and Temporal Memory Tracking}
\section{SpokenTrace: A Diagnostic Benchmark for Multi-Session Who-What-When Memory}
\label{sec:benchmark}

%We introduce \textbf{SpokenTrace}, a benchmark that evaluates whether speech agents can resolve speaker references, recall past information, and track how facts evolve over time across multi-session, multi-speaker conversations.

We introduce \textbf{SpokenTrace}, a benchmark that evaluates whether speech agents can recover \textit{Who} said \textit{What} and \textit{When} across multi-session, multi-speaker conversations by resolving speaker references, recalling past information, and tracking how facts evolve over time.

\subsection{Benchmark Design}
\label{sec:benchmark_design}

Each episode contains two historical dialogue sessions followed by a short current session containing a spoken query.
The histories contain recurring speakers, revisited topics, and facts that may be introduced, revised, or revisited across sessions.
Answering a query can therefore require identifying not only relevant content, but also whose statement it is, when it occurred, and which state of the information is being requested.

%To systematically evaluate these abilities and their combinations, we organize queries along two complementary annotation axes. The first dimension, \emph{memory task}, specifies the operation required by the query. The six history-dependent tasks are direct recall, temporal localization, latest-state retrieval, historical-state retrieval, change comparison, and multi-hop reasoning.The second, \emph{speaker-query type}, specifies how the query relates to speaker identity: it may be speaker-independent, refer to the current speaker through first-person expressions, refer to another participant by name, or ask the agent to identify a speaker from a fact or action. Crossing these dimensions tests whether agents can correctly resolve speaker references while performing different memory tasks. For example, ``When did I originally plan to leave?'' requires both identifying the current speaker in the dialogue history and recovering the departure plan before it was revised. 

To systematically evaluate these abilities and their combinations, we organize queries along two complementary annotation axes. The first dimension, \emph{memory task}, specifies the operation required by the query. The six history-dependent tasks are direct recall, temporal localization, latest-state retrieval, historical-state retrieval, change comparison, and multi-hop reasoning. The second, \emph{speaker-query type}, specifies how the query relates to speaker identity: it may be speaker-independent, refer to the current speaker through first-person expressions, refer to another participant by name, or ask the agent to identify a speaker from a fact or action. Together, these two axes disentangle memory reasoning from speaker grounding, while their combination tests whether agents can resolve speaker references correctly across different memory operations. For example, ``When did I originally plan to leave?'' requires both identifying the current speaker in the dialogue history and recovering the departure plan before it was revised.

We also include current-only queries, which can be answered from the current session, and missing-evidence queries, for which neither the historical sessions nor the current session provides supporting evidence. These controls test whether agents can answer using current information and avoid unsupported recall when evidence is absent.

Appendix~\ref{app:dataset-details} describes how the dataset is constructed and reports the number of queries in each task category, along with the sizes of the training, validation, and evaluation splits.

\subsection{Evaluation Protocol}
\label{sec:evaluation_protocol}

We evaluate systems at two levels: evidence retrieval and end-to-end
answering. The former measures recovery of annotated supporting
evidence, while the latter measures whether the system answers the
query correctly.

\paragraph{Evidence retrieval.}
The 1,200 evaluation episodes are partitioned into scenario-balanced
candidate pools, with an average of 4,331 candidate records per pool.
Each pool contains all historical records from its constituent
episodes.
Retrieval is evaluated using annotated memory records on the
1,106 queries that require historical evidence, with annotated
supporting utterances treated as positives; current-only and
missing-evidence queries are excluded.

We report Recall@1/3/5 and all-support EM@3/5.
Recall@$k$ measures the mean fraction of annotated supporting
utterances recovered per query among the top-$k$ results, while
all-support EM@$k$ measures the proportion of queries for which
all annotated supporting utterances appear in the top-$k$ results.
Formal metric definitions are provided in
Appendix~\ref{app:retrieval_metrics}.

\paragraph{End-to-end answering.}
End-to-end evaluation uses the same candidate-pool partition and
historical scope, with memory records produced by each system's
writer as described in Section~\ref{sec:experiments}.
All 1,200 evaluation queries are included in the end-to-end evaluation. We report strict accuracy using GPT-4.1-mini as an automatic judge~\citep{zheng2023judgingllmasajudgemtbenchchatbot}. Each response is assessed against the query, reference answer, and annotated evidence under task- and speaker-query-specific grading criteria, and is classified as \emph{correct}, \emph{partial}, or \emph{incorrect}. Only \emph{correct} responses count toward strict accuracy. The complete grading rubric and implementation details are provided in Appendix~\ref{app:e2e-evaluation}.

\section{Experiments}
\label{sec:experiments}

% We evaluate whether PERSIST can recover the evidence needed
% for conversational memory, how its native components support
% memory writing, and what additional cost it introduces.
% End-to-end evaluation tests the complete system, while
% component experiments examine retrieval, transcription,
% and speaker attribution separately.
%We evaluate whether PERSIST recovers the evidence needed for conversational memory, how well its native components support memory writing, and what additional cost it introduces. We measure the complete system end to end, then analyze retrieval, transcription, and speaker attribution separately.
We evaluate \textsc{PERSIST} from four perspectives: end-to-end conversational memory, Who-What-When retrieval, memory-writing quality, and full-duplex retrieval efficiency. We first evaluate the complete system on SpokenTrace, then analyze the retrieval and writing components separately.

\subsection{Training and implementation details}

\paragraph{System variants.}
We evaluate two implementations of \textsc{PERSIST}.
\textsc{PERSIST-Native} uses the backbone-native memory transcript branch, the speaker head, and persistent retriever described in Section~\ref{sec:system}.
\textsc{PERSIST-External} replaces only the transcript branch with VibeVoice-ASR, while speaker-grounding, retrieval, and dialogue components remain fixed.
% This provides a controlled comparison between native and plug-and-play memory writing while keeping the downstream memory pipeline fixed.
This comparison isolates the effect of native versus plug-and-play memory writing on downstream performance.

\paragraph{Native memory transcript branch.}
The transcript branch is initialized from the dialogue model by splitting after decoder layer 8 and copying the remaining 24 decoder layers into a dedicated memory branch.
Only branch-specific LoRA parameters, a lightweight token adapter, and the branch normalization are optimized, while the shared speech encoder and dialogue backbone remain frozen.
Training proceeds through timestamped ASR pretraining on GigaSpeech \citep{GigaSpeech} and CommonVoice \citep{commonvoice}, full-duplex conversational SFT on dialogues derived from Natural Questions and TriviaQA \citep{triviaqa}, and GRPO adaptation on the \textsc{SpokenTrace} training set using model-generated streaming history.

Training and inference use the same incremental transcription protocol.
Every 30 seconds, the branch receives up to 120 seconds of cached split-layer representations and recent committed transcripts, and generates only newly completed utterances with timestamps, without re-encoding the earlier audio.

\paragraph{Speaker grounding.}
For each timestamped transcript segment, a lightweight speaker head combines hidden states from layers 5, 7, and 8 of the frozen speech encoder through learned layer mixing and attentive statistics pooling. It produces a normalized, 4096-dimensional speaker embedding.
The speaker head is trained on speaker-disjoint CommonVoice segments.
These embeddings are matched online against a persistent speaker memory, which is initialized from registered speakers when available and updated as new observations arrive. Detailed speaker-head training and online assignment procedures are provided in Appendix~\ref{app:speaker-training}.

\paragraph{Retriever training.}
Training consists of two stages. Stage 1 learns general
audio-to-text retrieval using 196,929 prepared examples
from MS MARCO, QReCC, and TopiOCQA. Query speech is
synthesized with CosyVoice, with voices sampled from
a bank of 2,338 speakers. Stage 2 adapts the retriever to conversational memory
using the SpokenTrace training split. Detailed training hyperparameters are provided in
Appendix~\ref{app:training-details}.

The training pipeline uses a combination of
source-provided negatives, BM25 candidates,
neighboring conversational evidence, and dense
false positives mined during training.
For conversational adaptation, each query is contrasted
against all historical memories in its episode,
together with cross-rank negatives and explicitly
selected hard negatives, including memories with
similar or identical content but incorrect speakers or temporal states.

\subsection{End-to-end conversational memory}
\label{sec:e2e-experiments}
%We first test whether persistent Who-What-When memory improves end-to-end spoken dialogue across the memory demands defined by SpokenTrace  using the strict answer accuracy defined in Section~\ref{sec:evaluation_protocol}. We compare two configurations of PERSIST: \textsc{PERSIST-Native}, which uses the backbone-native transcription branch, and \textsc{PERSIST-External}, which replaces this branch with VibeVoice-ASR~\citep{peng2026vibevoiceasrtechnicalreport}. Both configurations use the same speaker-grounding module, retriever, and dialogue model. To compare retrieval methods under the same memory-writing configuration, we additionally replace the PERSIST retriever in the external configuration with BGE-large-en-v1.5 or BM25.
We first evaluate whether persistent Who-What-When memory improves end-to-end performance on SpokenTrace using the strict answer accuracy defined in Section~\ref{sec:evaluation_protocol}. 
We compare two configurations of \textsc{PERSIST}: \textsc{PERSIST-Native}, which uses the backbone-native transcription branch, and \textsc{PERSIST-External}, which replaces this branch with VibeVoice-ASR~\citep{peng2026vibevoiceasrtechnicalreport}. Both configurations share the same speaker-grounding module, retriever, and dialogue model. To compare retrieval methods under the same memory-writing configuration, we additionally replace the \textsc{PERSIST} retriever in the external configuration with BGE-large-en-v1.5 or BM25.

We also evaluate two reference conditions using the same
dialogue model.
\textit{Empty} answers using only the current-session
context, without retrieved historical records.
\textit{Gold} supplies three chronologically ordered memory
records containing the annotated supporting evidence whenever
retrieval is triggered.
These conditions assess performance without historical memory
and with oracle evidence, respectively.

\begin{table*}[h]
\centering
\setlength{\tabcolsep}{3pt}
\caption{Strict answer accuracy (\%) by memory task on
SpokenTrace. \textsc{PERSIST-External} (External), BGE, and BM25 use
VibeVoice-ASR transcripts with the PERSIST, BGE, and BM25
retrievers, respectively. \textsc{PERSIST-Native} (Native) uses the
backbone-native transcription branch and PERSIST retriever. Empty and Gold assess performance without historical memory
and with oracle evidence, respectively.}
\label{tab:e2e-memory-by-task}
\begin{tabular}{lrcccccc}
\toprule
Memory task & Queries & Gold & Empty & External & Native & BGE & BM25 \\
\midrule
Direct recall         & 194 & 96.39 &  1.55 & 85.57 & 64.95 & 82.99 & 80.93 \\
Historical state      & 197 & 99.49 &  1.02 & 87.31 & 72.08 & 49.75 & 58.88 \\
Latest state          & 200 & 97.00 &  0.00 & 82.00 & 65.00 & 42.00 & 47.00 \\
Change comparison     & 203 & 94.58 &  3.45 & 75.86 & 56.65 & 30.05 & 33.50 \\
Multi-hop composition & 133 & 96.99 &  0.00 & 75.19 & 61.65 & 25.56 & 33.08 \\
Temporal localization & 179 & 98.32 & 68.72 & 97.21 & 96.65 & 95.53 & 93.85 \\
Current-only control  &  40 & 97.50 & 97.50 & 97.50 & 97.50 & 97.50 & 97.50 \\
Missing evidence      &  54 &100.00 &100.00 & 96.30 &100.00 &100.00 &100.00 \\
\midrule
Overall               &1200 & 97.25 & 19.00 & 85.08 & 71.75 & 58.50 & 61.67 \\
\bottomrule
\end{tabular}
\end{table*}
Table~\ref{tab:e2e-memory-by-task} shows that
\textsc{PERSIST-External} achieves 85.08\% accuracy,
outperforming BM25 (61.67\%) and BGE (58.50\%)
with the same transcripts.
Gains are larger on state retrieval (historical and latest),
change comparison, and multi-hop composition than on
direct recall, for which content-based retrieval is
often sufficient.
\textsc{PERSIST-Native} reaches 71.75\%, highlighting
the impact of memory-writing quality.
Although \textit{Empty} reaches 68.72\% on temporal
localization, where answers lie in a small session-label
space, it achieves only 1.29\% across the other five
history-dependent tasks, compared with 96.87\% for
\textit{Gold}.
The external writer yields the highest end-to-end accuracy due to the much larger training corpus used in building VibeVoice-ASR, while the native configuration evaluates the fully backbone-integrated realization used for low-latency full-duplex deployment; both share the same speaker grounding and 3W retrieval formulation.

\subsection{Who-What-When Memory retrieval}
\label{sec:retrieval-experiments}
We follow the scenario-balanced retrieval protocol defined
in Section~\ref{sec:evaluation_protocol}.
We compare BM25~\citep{10.1561/1500000019},
BGE-large-en-v1.5~\citep{bge_embedding}, and PERSIST.
To assess the contribution of explicit temporal scoring,
we also evaluate the same trained PERSIST model with
its temporal score disabled at inference.

\begin{table}[h]
\caption{Memory retrieval results (\%) for BM25, BGE, and PERSIST
on history-dependent queries in SpokenTrace.
PERSIST w/o temporal score disables temporal scoring
at inference.
R@k denotes recall, and EM@k measures complete recovery
of the annotated supporting evidence.}
\label{tab:retrieval-overall}
\centering
\begin{tabular}{lrrrrr}
\toprule
Retriever & R@1 & R@3 & R@5 & EM@3 & EM@5 \\
\midrule
BM25
& 47.60 & 60.90 & 64.87 & 54.16 & 58.50 \\
BGE-large-en-v1.5
& 44.39 & 55.97 & 59.00 & 49.01 & 51.81 \\
PERSIST w/o temporal
& 62.84 & 85.08 & 91.18 & 80.20 & 87.79 \\
PERSIST
& \textbf{66.14} & \textbf{87.07} & \textbf{92.95}
& \textbf{82.10} & \textbf{89.51} \\
\bottomrule
\end{tabular}
\end{table}

Table~\ref{tab:retrieval-overall} shows that PERSIST
outperforms BM25 and BGE across all reported metrics,
achieving 82.10\% all-support EM@3.
Disabling temporal scoring reduces EM@3 to 80.20\%
and R@1 from 66.14\% to 62.84\%, indicating that the
explicit temporal component further improves retrieval.

\begin{table*}[h]
\centering
\caption{Retrieval results (\%) by memory task:
(a) Recall@1 on single-evidence tasks;
(b) recall and all-support exact match on two-evidence tasks.
``No temporal'' denotes PERSIST with temporal scoring
disabled at inference.
Bold indicates the best score for each task and metric.}
\label{tab:retrieval-tasks}
\fontsize{8}{8.5}\selectfont
\setlength{\tabcolsep}{2pt}

\begin{minipage}[t]{0.56\textwidth}
\vspace{0pt}
\centering
\textbf{(a) Single-evidence tasks: R@1.}
\par\smallskip

\begin{tabular}{@{}lrrrrr@{}}
\toprule
Task & Queries & BM25 & BGE & No temporal & PERSIST \\
\midrule
Direct recall
& 194 & 80.93 & 79.38 & 83.51 & \textbf{84.02} \\
Latest state
& 200 & 24.50 & 23.00 & 72.50 & \textbf{74.00} \\
Historical state
& 197 & 36.55 & 28.43 & 73.10 & \textbf{80.20} \\
Temporal localization
& 179 & 86.03 & \textbf{88.27} & 78.77 & 83.80 \\
\bottomrule
\end{tabular}

\par\vspace{6pt}
{\normalsize\raggedright
\textbf{Results by memory task.}
For single-evidence tasks, Recall@1 measures whether
the supporting record is ranked first.
For two-evidence tasks, Recall@3/5 and all-support
EM@3/5 measure partial and complete evidence recovery,
respectively.
\par}

\end{minipage}
\hfill
\begin{minipage}[t]{0.42\textwidth}
\vspace{0pt}
\centering
\textbf{(b) Two-evidence tasks.}
\par\smallskip

\begin{tabular}{@{}lrrrr@{}}
\toprule
Retriever & R@3 & R@5 & EM@3 & EM@5 \\
\midrule
\multicolumn{5}{@{}l}{\textit{Change comparison ($n=203$)}} \\
BM25
& 32.76 & 36.95 & 24.63 & 29.06 \\
BGE
& 27.34 & 31.53 & 17.73 & 21.67 \\
No temporal
& \textbf{78.08} & 88.42 & 63.55 & 80.30 \\
PERSIST
& 77.34 & \textbf{88.92} & \textbf{64.04} & \textbf{81.28} \\
\midrule
\multicolumn{5}{@{}l}{\textit{Multi-hop composition ($n=133$)}} \\
BM25
& 56.39 & 59.02 & 12.78 & 18.05 \\
BGE
& 45.49 & 46.99 & 2.26 & 2.26 \\
No temporal
& 53.01 & 63.91 & 34.59 & 48.12 \\
PERSIST
& \textbf{58.65} & \textbf{73.31}
& \textbf{37.59} & \textbf{56.39} \\
\bottomrule
\end{tabular}
\end{minipage}

\end{table*}
\FloatBarrier
The gains from temporal scoring vary across tasks.
Recall@1 improves only slightly on direct recall
(83.51\% to 84.02\%), but more substantially on
historical-state retrieval (73.10\% to 80.20\%).
Temporal localization also improves from 78.77\% to
83.80\%, although BGE remains stronger at 88.27\%.
This task asks when an event occurred; retrieving
the event by its content and then reading its timestamp
can be sufficient.

On two-evidence tasks, PERSIST achieves higher
all-support EM@3 than BM25 and BGE, reaching 64.04\%
on change comparison and 37.59\% on multi-hop composition.
Temporal scoring provides a larger EM@3 improvement
on multi-hop composition (34.59\% to 37.59\%) than
on change comparison (63.55\% to 64.04\%).

\par\noindent
\begin{minipage}{\linewidth}

\noindent
\begin{minipage}[t]{0.51\linewidth}
\vspace{0pt}
\textbf{Results by speaker-query type.}\par
PERSIST achieves the highest EM@3 across all four
query types, with particularly large gains on
self-to-fact queries
(Table~\ref{tab:retrieval-speakers}).

\par\smallskip
To test the contribution of acoustic speaker identity,
we construct 397 gold--counterfactual pairs across
290 first-person queries.
Each counterfactual replaces only the speaker token
with another speaker's token from the same episode.
PERSIST assigns a higher score to the record with
the correct speaker in 96.47\% of pairs,
demonstrating sensitivity to speaker identity with
text, names, and timestamps held fixed.
Adding these counterfactuals to the candidate pools
reduces R@1 by only 1.38 percentage points,
from 60.00\% to 58.62\%.\end{minipage}
\hfill
\begin{minipage}[t]{0.47\linewidth}
\vspace{0pt}

\begingroup
\setlength{\abovecaptionskip}{0pt}
\captionof{table}{Speaker-aware retrieval (\%). \\
Top: EM@3 by query type. \\ 
Bottom: PERSIST's counterfactual results.}
\label{tab:retrieval-speakers}
\endgroup

\centering
\small
\setlength{\tabcolsep}{4pt}
\renewcommand{\arraystretch}{1.05}
\begin{tabular}{@{}lrrr@{}}
\toprule
Query type & BM25 & BGE & PERSIST \\
\midrule
Self-to-fact
& 25.52 & 17.24 & \textbf{79.66} \\
Person-to-fact
& 68.47 & 69.15 & \textbf{88.47} \\
Fact-to-person
& 74.34 & 68.30 & \textbf{75.85} \\
Speaker-independent
& 49.22 & 41.80 & \textbf{83.98} \\
\midrule
\multicolumn{3}{@{}l}{\textit{Counterfactual test}}
& \textit{Score} \\
\multicolumn{3}{@{}l}{Gold preferred over counterfactual}
& 96.47 \\
\multicolumn{3}{@{}l}{R@1: original candidates}
& 60.00 \\
\multicolumn{3}{@{}l}{R@1: with counterfactuals}
& 58.62 \\
\bottomrule
\end{tabular}
\end{minipage}

\par\medskip
\noindent
Together with the temporal-score ablation
(Table~\ref{tab:retrieval-tasks}), these results provide
complementary evidence for the two components:
temporal scoring improves historical-state retrieval,
while acoustic speaker identity influences ranking
even when textual and temporal cues are held fixed.

\end{minipage}
\par\smallskip

\subsection{Memory Writing}
\label{sec:writing-experiments}

We evaluate memory writing on held-out full-duplex conversational SFT data in the single-speaker setting and on \textsc{Spokentrace} in the multi-speaker setting.
We report WER and temporal IoU in both settings, cpWER for multi-speaker transcription, and Spk-W Acc, Pair-F1, and New-Spk F1 for speaker grounding.

\begin{table}[!htbp]
\caption{
Memory-writing quality on single- and multi-speaker speech.
We evaluate transcription and temporal alignment in both settings, and additionally evaluate
speaker attribution and online identity assignment in the multi-speaker setting.
}
\label{tab:memory-writing-results}
\centering
\setlength{\tabcolsep}{1.8pt}
\begin{tabular}{lcccccccc}
\toprule
& \multicolumn{2}{c}{Single-speaker}
& \multicolumn{6}{c}{Multi-speaker} \\
\cmidrule(lr){2-3}\cmidrule(lr){4-9}
Model
& WER$\downarrow$ & IoU$\uparrow$
& WER$\downarrow$ & cpWER$\downarrow$ & IoU$\uparrow$ 
& Spk-W Acc$\uparrow$ & Pair-F1$\uparrow$ & New-Spk F1$\uparrow$ \\
\midrule
VibeVoice-ASR   & 2.4\% & 98.9\% &  5.3\% &  6.8\% & 98.5\% & 99.7\% & 99.6\% & 97.8\% \\
Native (Pre-trained) & 38.4\% & 57.6\% & 51.9\% & 87.5\% & 62.3\% & 60.5\% & 53.3\% & 65.7\% \\
\quad + SFT          & 11.7\% & 81.8\% & 35.1\% & 63.7\% & 71.8\% & 73.1\% & 73.5\% & 75.8\% \\
\quad + GRPO         &  6.8\% & 87.5\% & 17.4\% & 41.6\% & 79.4\%  & 80.4\% & 81.8\% & 83.3\% \\
\bottomrule
\end{tabular}
\end{table}

Table~\ref{tab:memory-writing-results} shows that staged
training reduces the native writer's WER from 38.4\%
to 6.8\% on single-speaker speech and from 51.9\%
to 17.4\% on multi-speaker speech.
Beyond SFT, GRPO further improves multi-speaker cpWER
from 63.7\% to 41.6\% and temporal IoU from 71.8\%
to 79.4\%, benefiting both speaker-attributed
transcription and temporal alignment.

Nevertheless, the native writer's multi-speaker cpWER
remains substantially higher than its WER
(41.6\% versus 17.4\%), indicating persistent
speaker-attribution errors.
VibeVoice-ASR achieves lower cpWER (6.8\%) and higher
temporal IoU (98.5\%) and Spk-W Acc (99.7\%).
This writing-quality gap is consistent with the
end-to-end advantage of \textsc{PERSIST-External} over
\textsc{PERSIST-Native} (Table~\ref{tab:e2e-memory-by-task}).

\subsection{Full-Duplex Retrieval Efficiency}
\label{sec:efficiency-experiments}

\noindent
\begin{minipage}[t]{0.58\linewidth}
\vspace{0pt}
Finally, we test whether \textsc{PERSIST} can meet the latency requirements of real-time full-duplex interaction.
We profile 1,200 SpokenTrace queries on an Ascend 910B3 NPU
(32 CPU threads), averaging the middle 1,000 requests.
Retrieval is timed through top-3 return, excluding live dialogue;
larger memory banks repeat caption vectors.
By reusing speech representations, PERSIST reduces retrieval latency
at 4,331 candidates from 578.42 ms (waveform recomputation)
to 7.03 ms, a 98.8\% reduction (Table~\ref{tab:efficiency}).
\end{minipage}\hfill
\begin{minipage}[t]{0.39\linewidth}
\vspace{0pt}
\centering

\makeatletter
\def\@captype{table}
\makeatother
\setlength{\abovecaptionskip}{0pt}
\setlength{\belowcaptionskip}{3pt}
\caption{Mean retrieval latency (ms).}
\label{tab:efficiency}

\small
\renewcommand{\arraystretch}{1.25}
\setlength{\tabcolsep}{3pt}
\begin{tabular}{@{}lr@{}}
\toprule
Method & Latency (ms) \\
\midrule
BM25 (text-ready)      & 15.26 \\
BGE-large (text-ready) & 27.94 \\
Recompute from WAV    & 578.42 \\
PERSIST (native, buffered) & \textbf{7.03} \\
\bottomrule
\end{tabular}
\end{minipage}
\par

\section{Conclusion}
In this work, we addressed the challenge of Who-What-When memory across sessions for spoken dialogue: retrieving not only what was said, but also who said it and which temporal state the query refers to. We presented \textsc{PERSIST}, a persistent memory architecture that stores readable conversational events and retrieves them with 3W joint scoring over semantic content, acoustic speaker identity, and temporal state. For real-time full-duplex interaction, \textsc{PERSIST} reuses intermediate dialogue-backbone representations to avoid redundant query-audio encoding, reducing retrieval latency from 578.42 ms to 7.03 ms. We also introduced SpokenTrace, a diagnostic benchmark that factorizes evaluation along memory tasks and speaker-query types. On SpokenTrace, \textsc{PERSIST} achieves 85.08\% end-to-end task accuracy and improves all-support EM@3 to 82.10\%. We hope \textsc{PERSIST} and SpokenTrace provide a foundation for persistent, speaker-aware, and temporally grounded memory in real-time spoken agents.

%In this work, we addressed the key challenge of long-term memory in dynamic multi-speaker spoken dialogue: maintaining joint speaker attribution and temporal state tracking over multi-session interactions. We presented PERSIST, a persistent memory architecture featuring a 3W joint scoring mechanism that unifies semantic content, acoustic speaker embeddings, and temporal representations for precise memory retrieval. By reusing intermediate representations from the full-duplex dialogue backbone, PERSIST eliminates redundant speech re-encoding and achieves an ultra-low retrieval latency of 7.03 ms. Furthermore, we introduced SpokenTrace, the first diagnostic benchmark specifically designed to evaluate persistent conversational memory across orthogonal memory tasks and speaker-query types. Extensive evaluations demonstrate that PERSIST significantly outperforms existing text-based and speech memory baselines, elevating end-to-end task accuracy to 85.08\% and all-support EM@3 to 82.10\%. We hope PERSIST and SpokenTrace serve as a foundation for future research on long-term, low-latency, and speaker-aware memory in real-time conversational agents.

\clearpage
\subsection*{AI use statement}

Generative AI tools were used to help organize the paper's narrative, edit language, draft LaTeX, and discuss figure layouts. The authors reviewed the generated text against the implementation and primary literature, and will verify all experimental claims, citations, code, and artifacts before submission. The tools were not treated as an authority for scientific results. The authors take responsibility for the final content of this work, including text, claims, and artifacts produced with the aid of generative AI.

\subsection*{Ethics statement}

Persistent conversational memory introduces privacy and consent risks beyond those of a session-only assistant. Any release or deployment should make memory collection visible, restrict records to the correct user and application, support inspection and deletion, and avoid interpreting a voice embedding as real-world identity. Dataset licenses, speaker consent, demographic coverage, and access controls will be documented with the final experiments.

\subsection*{Reproducibility statement}

The final paper will specify the streaming schedule, layer indices, parameter accounting, data construction, speaker splits, hard-negative mining, trigger labels, all evaluation prompts, and the serving hardware used for latency and memory measurements. Code and configuration files will be included in anonymous supplementary material where policy permits.

\bibliography{references}
\bibliographystyle{iclr2027_conference}

\appendix
\section{Related Work}
\label{sec:related}

\paragraph{Full-duplex speech models.}
Full-duplex systems model listening and speaking concurrently. Moshi jointly models user and system streams through audio tokens~\citep{defossez2024moshi}, while SALMONN-Omni connects a streaming speech encoder and synthesizer to one codec-free language backbone~\citep{yu2025salmonnomni}. Our work studies what happens after this interaction model exceeds a finite session: we preserve its duplex path and expose memory-relevant information from the same hidden states.

\paragraph{Streaming context and persistence.}
StreamingLLM preserves fluency under a bounded sliding KV cache by retaining attention-sink tokens~\citep{xiao2024streamingllm}. It motivates our treatment of the rapidly growing full-duplex context, but it solves a different problem from persistent memory: bounded streaming state neither restores semantically evicted evidence nor survives a new session. Our explicit caption store complements rather than replaces bounded-cache methods.

\paragraph{Explicit and lifelong memory.}
Our working KV cache is implicit and short-term, whereas the caption bank is explicit, online-updated, and cross-session. PERSIST studies this distinction through online speech-memory writing and retrieval. Our question is how a spoken agent can write, attribute, retrieve, and use explicit memory while preserving its live interaction behavior.

\paragraph{Retrieval with shared backbones.}
An explicit non-parametric index provides updatable evidence and provenance. Self-RAG learns reflection tokens that let one model retrieve and critique its generations~\citep{asai2024selfrag}, while WavRAG studies native-audio retrieval for external knowledge~\citep{chen2025wavrag}. We use these ideas as references for control-token routing and spoken-query encoding, but evaluate a different problem: retrieving attributed events from the agent's own conversational history while reusing the live audio path.

\paragraph{Benchmarks for conversational memory.}
Existing speech and retrieval benchmarks typically isolate one capability: transcription, speaker recognition, text retrieval, or answer generation from a completed turn. They therefore do not test whether a system writes memory continuously, retrieves after a session reset, tracks changing facts, composes evidence across events, grounds different kinds of speaker queries, or avoids retrieving when evidence is current or absent. SpokenTrace varies these memory tasks and speaker relations independently over long-horizon event timelines. Its purpose is not to replace specialized metrics, but to expose the end-to-end failures that arise when they are composed in a full-duplex agent.

\section{SpokenTrace Construction and Dataset Details}
\label{app:dataset-details}

\subsection{Benchmark Construction Pipeline}
\label{app:benchmark-construction}

\paragraph{Scenario coverage.}
\textsc{SpokenTrace} covers 24 conversational subscenarios grouped into three broad domains: work and project collaboration, media and entertainment, and shared living and social planning.
Each domain contains eight subscenarios spanning different conversational settings and topics, as summarized in Table~\ref{tab:spokentrace-scenarios}.

\begin{table}[t]
\centering
\small
\caption{Scenario coverage in \textsc{SpokenTrace}. Each domain contains eight subscenarios.}
\label{tab:spokentrace-scenarios}
\begin{tabular}{lll}
\toprule
\textbf{Work / Project} &
\textbf{Media / Entertainment} &
\textbf{Shared Living / Social} \\
\midrule
Speech/audio model research      & Movie night              & Shared apartment chores \\
Dataset collection study         & TV-series club           & Grocery/meal planning \\
Software/model release           & Music listening          & Restaurant choice \\
Paper/experiment coordination    & Video review             & Weekend activity \\
Prototype/demo preparation       & Documentary discussion   & Travel itinerary \\
Infrastructure migration         & Podcast/video club       & Household shopping \\
User study                       & Playlist curation        & Social event \\
Benchmark/challenge preparation  & Home screening           & Errand coordination \\
\bottomrule
\end{tabular}
\end{table}

\paragraph{Plan-first episode specification.}
\textsc{SpokenTrace} is constructed using a plan-first generation pipeline in which the supervision for each episode is defined before the conversational dialogue is realized.
A programmatic scheduler first selects the scenario, subscenario, memory task, semantic relation, and task-specific control variables.
The corresponding task contract then defines a structured world state containing the benchmark-critical facts, their temporal and speaker relations, the gold answer, the supporting facts, and the query logic.
For tasks involving updates or comparisons, the world state additionally specifies the relation between earlier and later fact states.
This separation ensures that answers and supporting evidence are determined from the structured task specification rather than inferred post hoc from generated dialogue.
Only benchmark-critical information is fixed at this stage; the dialogue generator is free to introduce noncritical conversational details for naturalness.

\paragraph{Historical dialogue realization.}
Each episode contains two historical sessions followed by a shorter current session.
A programmatic blueprint assigns benchmark-critical facts to their designated historical sessions and distributes them across shorter conversational segments for generation.
Each historical segment is generated from only its assigned facts and recent conversational context. Construction-only annotations, such as fact IDs and internal metadata, are not exposed to the dialogue generator and are maintained programmatically instead.
Benchmark-critical facts are associated with canonical surface forms that are required to occur in their designated session, enabling unambiguous alignment between structured supporting facts and realized utterances.
The remaining turns are generated freely to provide surrounding discussion, clarifications, and distractors.
For tasks involving temporal state changes, later-state information is not exposed when the earlier historical session is generated, preventing future information from leaking into the earlier dialogue.

\paragraph{Current-session construction.}
The current session is generated after the two historical sessions.
For history-dependent tasks, the dialogue generator receives only the information needed to produce the pre-query conversation and is not given the historical gold answer or its supporting evidence.
After the pre-query dialogue is generated, the final benchmark query is appended programmatically according to the pre-specified task and target speaker.
This keeps the query and query speaker fixed while preventing the current dialogue from leaking the historical answer.

\paragraph{Generation model routing.}
Rather than using a single language model for all stages, we use stage-specific model routing, fixed after pilot comparisons of generation quality for each component of the construction pipeline.
GPT-5-mini-2025-08-07 is used to instantiate structured world states and to generate and repair benchmark questions.
Conversational dialogue is realized using a mixture of Qwen3-30B-A3B-Instruct-2507 and GPT-5-mini-2025-08-07.
For each episode, one dialogue model is selected and used throughout the episode, with sampling probabilities of $0.8$ for Qwen3-30B-A3B-Instruct-2507 and $0.2$ for GPT-5-mini-2025-08-07.
The generated dialogue follows the prescribed session structure and benchmark facts, while the surrounding conversational content remains freely realized by the selected model.

\subsection{Memory Tasks and Speaker-Query Types}
\label{app:task-taxonomy}

Each query in \textsc{SpokenTrace} is annotated along two complementary axes:
a \emph{memory task}, which specifies what information must be recovered from the conversation history, and a \emph{speaker-query type}, which specifies how the query refers to conversational participants.
This factorization separates the reasoning required over conversational history from the way speaker identity enters the query.
The two axes are defined separately for analysis, although their empirical
distributions are not assumed to be statistically independent.

\paragraph{Memory tasks.}
We group queries into eight memory tasks according to the evidence required to
answer them:

\begin{table}[!htbp]
\caption{
Definitions of the eight memory tasks in \textsc{SpokenTrace}.
}
\label{tab:appendix-memory-tasks}
\centering
\setlength{\tabcolsep}{5pt}
\begin{tabular}{lp{0.70\linewidth}}
\toprule
Task & Definition \\
\midrule
Direct recall
& Recover a fact stated in the historical conversation. \\

Temporal localization
& Identify when, or in which historical session, a referenced fact or discussion occurred. \\

Latest state
& Recover the most recent value of a fact after an earlier value has been updated or superseded. \\

Historical state
& Recover an earlier value of a fact despite a later update being present in the history. \\

Change comparison
& Recover both an earlier and a later state and describe how the fact changed between them. \\

Multi-hop
& Combine information from multiple supporting records to derive the answer. \\

Current-only
& Answer from information introduced in the current session rather than from persistent history. \\

Missing evidence
& Recognize that the requested information is not supported by either the historical or current conversation. \\
\bottomrule
\end{tabular}
\end{table}

These task definitions are enforced during benchmark construction by specifying the required supporting evidence before dialogue generation.
Current-only and missing-evidence queries are included as controls: the former should be answered from the current context without relying on historical memory, whereas the latter contains no supporting evidence and requires the system to recognize that the requested information is unavailable.

\paragraph{Speaker-query types.}
Independently of the memory operation, queries are categorized by how they refer to speakers:

\begin{table}[!htbp]
\caption{Speaker-query types in \textsc{SpokenTrace}.}
\label{tab:appendix-speaker-query-types}
\centering
\setlength{\tabcolsep}{5pt}
\begin{tabular}{lp{0.70\linewidth}}
\toprule
Type & Definition \\
\midrule
Speaker-independent
& The answer does not depend on identifying a particular participant. \\

Self-to-Fact
& The current speaker refers to themself, requiring the system to recover that speaker's historical information. \\

Person-to-Fact
& The query explicitly refers to a person and asks for information associated with that person. \\

Fact-to-Person
& The query provides a fact or action and asks which participant it should be attributed to. \\
\bottomrule
\end{tabular}
\end{table}

Among these categories, Self-to-Fact queries directly require cross-session acoustic speaker continuity: the system must associate the current speaker with that same participant's historical speech.
Person-to-Fact and Fact-to-Person queries use speaker references expressed in the conversation text, while Speaker-independent queries do not impose a speaker-specific constraint.
Together, the two axes allow the benchmark to distinguish failures of conversational memory from failures of speaker grounding.

\subsection{Dataset Splits and Statistics}

\textsc{SpokenTrace} is divided into training, validation, and evaluation splits. The training split contains episodes not used for validation or evaluation.
For evaluation, we select one query from each of 1,200 episodes, yielding 1,200 evaluation queries. Table~\ref{tab:spokentrace_splits} summarizes the dataset sizes.

\begin{table}[t]
\centering
\small
\caption{Dataset splits of \textsc{SpokenTrace}. Episode counts denote the number
of distinct multi-session conversational histories represented in each split.}
\label{tab:spokentrace_splits}
\begin{tabular}{lrr}
\toprule
\textbf{Split} & \textbf{Episodes} & \textbf{Queries} \\
\midrule
Training    & 9,119 & 27,985 \\
Validation  & 1,013 & 3,185  \\
Evaluation  & 1,200 & 1,200  \\
\bottomrule
\end{tabular}
\end{table}

The evaluation set is constructed to cover the full range of memory demands and speaker references in the benchmark.
As shown in Table~\ref{tab:spokentrace_design}, the four speaker-query types are exactly balanced with 300 queries each.
The memory-task distribution covers all six history-dependent task families, with current-only and missing-evidence queries included as smaller control subsets.
Overall, the evaluation set contains 1,200 queries across all eight memory tasks and four speaker-query types.

\begin{table}[t]
\centering
\small
\caption{Task and speaker-query distributions in the \textsc{SpokenTrace} evaluation set.}
\label{tab:spokentrace_design}
\begin{tabular}{lrlr}
\toprule
\multicolumn{2}{c}{\textbf{Memory task}} &
\multicolumn{2}{c}{\textbf{Speaker-query type}} \\
\cmidrule(lr){1-2}\cmidrule(lr){3-4}
\textbf{Type} & \textbf{Queries} &
\textbf{Type} & \textbf{Queries} \\
\midrule
Direct recall         & 194 & Speaker-independent & 300 \\
Temporal localization & 179 & Self-to-Fact        & 300 \\
Latest state          & 200 & Person-to-Fact      & 300 \\
Historical state      & 197 & Fact-to-Person      & 300 \\
Change comparison     & 203 &                     &     \\
Multi-hop             & 133 &                     &     \\
Current-only          & 40  &                     &     \\
Missing evidence      & 54  &                     &     \\
\bottomrule
\end{tabular}
\end{table}

\subsection{Speech Synthesis and Voice Assignment}

After constructing the textual episodes, we synthesize all dialogue utterances with CosyVoice2~\citep{du2024cosyvoice2scalablestreaming}.
Speaker references are drawn from LibriSpeech\citep{librispeech}.
For each reference speaker, we use one source utterance together with its transcript as the audio--text prompt for speech synthesis.

For each episode, the conversational participants are assigned distinct reference speakers.
The participant--speaker mapping is fixed throughout the episode: the same participant uses the same reference speaker in both historical sessions and in the current session, while different participants use different speakers.
This ensures that speaker identity remains acoustically consistent across sessions, allowing speaker-dependent memory queries to be resolved from the speech signal rather than from explicit textual speaker labels.

\subsection{Quality Control and Audits}

Each generated episode undergoes both deterministic validation and semantic auditing before being included in the benchmark.
The deterministic checks verify structural and annotation consistency, including session and query structure, realization of the prescribed benchmark facts, and consistency between each query, its gold answer, and its annotated supporting evidence.
They also check for unintended information leakage, such as construction-only metadata appearing in the dialogue, future-state information appearing in an earlier session, or the historical gold answer being exposed in the current pre-query context.

We then apply a task-specific semantic audit to the completed dialogue.
This audit verifies that the generated conversation actually realizes the intended memory task and that the gold answer is supported by the prescribed evidence.
For state-tracking tasks, it checks that earlier and later values have the intended temporal relation; for multi-hop queries, it verifies that the required supporting facts must genuinely be combined rather than allowing either one to answer the query alone.
For current-only controls, the answer must be available from the current session while historical context contains only a stale alternative, whereas missing-evidence queries must remain unsupported by both historical and current dialogue.
Speaker-dependent queries are also checked to ensure that the answer is uniquely determined only when the relevant cross-session speaker identity is available.
Episodes that fail a hard deterministic or semantic check are regenerated or excluded.
\section{PERSIST Implementation Details}

\subsection{Memory Writer Implementations}
\label{app:memory-writers}

\paragraph{Backbone-native writer.}
The backbone-native writer converts the ongoing speech stream into readable memory records without introducing a separate ASR encoder.
It reuses intermediate speech representations already computed by
SALMONN-Omni during regular full-duplex inference and applies an ASR-specific route with LoRA adapters over the final 24 decoder layers to produce utterance-level transcripts.
The writer operates on cached backbone representations and commits only newly completed utterances, avoiding repeated encoding of previously processed audio.
Each committed transcript is associated with its session and temporal span and is subsequently passed to the shared speaker-grounding and memory-encoding pipeline.

\paragraph{External writer.}
For the external configuration, we replace the native transcription route with VibeVoice-ASR~\citep{peng2026vibevoiceasrtechnicalreport}.
The resulting utterance transcripts follow the same memory-record interface and are processed by the same speaker-grounding, persistent storage, and retrieval components as in the backbone-native configuration.
Thus, the two implementations differ only in how conversational speech is transcribed into readable records, while the downstream memory pipeline remains unchanged.

\subsection{Speaker Grounding and Persistent Identity}
\label{app:speaker-grounding}

\paragraph{Speaker representation.}
For each timestamped utterance segment, we reuse hidden states from layers 5, 7, and 8 of the SALMONN-Omni speech encoder rather than introducing a separate speaker encoder.
A learned layer mixture followed by attentive statistics pooling produces a 4096-dimensional L2-normalized speaker embedding.
Cosine similarity between these embeddings is used for online speaker matching. The same speaker representation is used to associate stored utterances with speaker identities and to provide speaker information for memory retrieval.

\paragraph{Persistent speaker assignment.}
We maintain a running centroid for each conversation-level speaker identity.
For a newly observed utterance, its speaker embedding is compared with all
existing centroids; it is assigned to the closest speaker when the maximum
cosine similarity exceeds a threshold selected on the validation set, and a new speaker identity is created otherwise.
When an existing speaker is selected, its centroid is updated by a duration-weighted mean of the previous centroid and the new embedding, followed by normalization.
These identities persist across successive ASR windows and memory records.
\section{Training Details}
\label{app:training-details}

\subsection{Native Memory Writer Training}
\label{app:native-writer-training}

The native memory writer is trained in three stages while the shared SALMONN-Omni speech encoder and dialogue backbone remain frozen.
Timestamped ASR pretraining uses 50,517 GigaSpeech and 50,712 Common Voice examples (101,229 examples; approximately 3,374.3 hours).
Conversational SFT then uses 219,628 multi-turn conversations constructed from Natural Questions and TriviaQA, with a 120-s input context.
We train this stage for 10,000 steps with a global batch size of 32 and decay the learning rate from $5\times10^{-6}$ to $1\times10^{-6}$.
Finally, GRPO is applied to 41,299 windows from the \textsc{SpokenTrace}
training split (approximately 1,376.6 hours).

For GRPO, we define a transcription reward
\[
r_{\mathrm{text}}
=
1-\frac{\min(\mathrm{WER}_{\mathrm{session}},\,1.5)}{1.5},
\]
Here, $\mathrm{WER}_{\mathrm{session}}$ is computed by concatenating the reference and generated transcripts from all windows up to the sampled target boundary.
No cross-window deduplication is applied when computing this reward.
The format reward $r_{\mathrm{format}}$ evaluates whether the output can be parsed into valid timestamped utterances with non-empty text and without abnormal special-token outputs.
The temporal reward $r_{\mathrm{time}}$ measures timestamp IoU after matching predicted and reference utterances by text, allowing one-to-many and many-to-one matches.

We use a short curriculum for temporal optimization:
\[
r =
\begin{cases}
0.75\,r_{\mathrm{text}} + 0.10\,r_{\mathrm{format}},
& k < 100,\\[2mm]
0.75\,r_{\mathrm{text}} + 0.10\,r_{\mathrm{format}}
+ 0.15\,r_{\mathrm{time}},
& k \geq 100,
\end{cases}
\]
where $k$ denotes the GRPO optimization step.
The final objective additionally retains supervised reference-token learning,
\[
\mathcal{L}
=
\mathcal{L}_{\mathrm{GRPO}}
+
0.2\,\mathcal{L}_{\mathrm{CE}}.
\]

For each GRPO target, we sample four responses with temperature 1.0,
top-$p=0.95$, and a maximum generation length of 256 tokens.
Training uses a global batch size of 16 for up to 3,000 steps, with the
learning rate decayed from $5\times10^{-7}$ to $1\times10^{-7}$.
We use a clipping coefficient of $\epsilon=0.2$ and no additional KL penalty
or reference-model regularization.
Only the ASR-specific LoRA and memory-writing branch parameters are updated.

\subsection{Speaker Head Training}
\label{app:speaker-training}

The speaker head is trained with the SALMONN-Omni speech encoder frozen.
Training uses speaker-balanced batches of speech segments and jointly optimizes
an additive angular-margin classification loss and a supervised contrastive loss,

\[
\mathcal{L}_{\mathrm{spk}}
=
\mathcal{L}_{\mathrm{AAM}}
+
0.1\,\mathcal{L}_{\mathrm{SupCon}}.
\]

Only the learned layer mixture and speaker-head parameters are updated.
The operating threshold used for online speaker assignment is selected on the validation set and fixed before downstream evaluation.

\subsection{Retriever Training}
\label{app:retriver-training}

\begin{table}[h]
\centering
\small
\caption{Retriever training hyperparameters.}
\begin{tabular}{lcc}
\toprule
Hyperparameter & Stage 1 & Stage 2 \\
\midrule
Optimizer & AdamW & AdamW \\
Retrieval-head learning rate & $3\times10^{-4}$ & $10^{-4}$ \\
Retrieval-LoRA learning rate & $10^{-4}$ & $2\times10^{-5}$ \\
Weight decay & $10^{-4}$ & $10^{-4}$ \\
Contrastive temperature & $0.05$ & $0.05$ \\
Accelerator ranks & 8 & 8 \\
Batch size per rank & 8 queries & 4 episodes \\
Global batch size & 64 queries & 32 episodes \\
Training epochs & 10 & 5 \\
Checkpoint selection & Val.\ R@3 & Val.\ EM@5 \\
\bottomrule
\end{tabular}
\end{table}
\section{Evaluation Details}

\subsection{End-to-End Answer Evaluation}
\label{app:e2e-evaluation}

We evaluate end-to-end responses with GPT-4.1-mini-2025-04-14.
For each query, the judge receives the question, reference answer, memory-task and speaker-query annotations, gold answer components, annotated supporting utterances, and the model response.
The judge assigns one of three labels: \emph{correct}, \emph{partial}, or \emph{incorrect}.
A response is marked correct only when it provides all answer slots requested by the question with the correct factual content, speaker attribution, and temporal interpretation.
Partial responses contain some correct required information but omit a required slot or mix a correct answer with a distinct incorrect alternative.
Responses with an incorrect core fact, speaker, temporal state, or session, as well as unjustified abstention on answerable queries, are marked incorrect.

Grading follows the semantics of each memory task rather than requiring lexical agreement with the reference.
In particular, latest-state and historical-state queries require only the requested state, change-comparison queries require both states when the change itself is requested, temporal-localization queries require the correct session or order, and multi-hop queries require all answer slots explicitly requested by the question.
For missing-evidence queries, an explicit statement that the requested information is unavailable is considered correct, whereas a specific unsupported guess is incorrect.
For speaker-conditioned queries, only the speaker- or fact-specific answer slots requested by the question are required; contextual facts used only to identify the target are not additionally required.

We apply a small set of deterministic safeguards before final scoring.
Normalized exact matches to an accepted reference answer and clear abstentions for missing-evidence queries are accepted directly, while malformed concatenations of otherwise distinct required answer slots are marked partial.
Exact repetition of the same answer is collapsed before judging.
These safeguards are deliberately restricted to unambiguous cases, with all remaining responses evaluated semantically by the judge.

Our primary metric is strict accuracy, for which only \emph{correct} responses receive credit; \emph{partial} and \emph{incorrect} responses both count as zero.
A diagnostic soft score assigns $1.0$, $0.5$, and $0.0$ to the three labels, respectively.
Judge inference uses temperature $0$ with a fixed seed and structured JSON
output.
API or parsing failures are reported as unjudged rather than silently counted as incorrect.

\subsection{Memory-Retrieval Evaluation}
\label{app:retrieval_metrics}

Retrieval metrics are computed over the evaluation queries that require historical evidence.
Let $\mathcal{Q}_{\mathrm{ret}}$ denote this query set, $G_q$ the set of annotated supporting utterances for query $q$, and $T_q^k$ the top-$k$ retrieved records.
We compute

\[
\operatorname{Recall@}k
=
\frac{1}{|\mathcal{Q}_{\mathrm{ret}}|}
\sum_{q\in\mathcal{Q}_{\mathrm{ret}}}
\frac{|T_q^k \cap G_q|}{|G_q|},
\]

and

\[
\operatorname{EM@}k
=
\frac{1}{|\mathcal{Q}_{\mathrm{ret}}|}
\sum_{q\in\mathcal{Q}_{\mathrm{ret}}}
\mathbf{1}\!\left[G_q \subseteq T_q^k\right].
\]

We report Recall@1/3/5 and all-support EM@3/5.
Recall@k measures the fraction of annotated supporting evidence recovered, whereas EM@k counts a query as correct only when all of its supporting utterances appear in the top-$k$ results.
Consequently, for queries requiring two supporting utterances, Recall@1 has a maximum value of $50\%$, while EM@k directly measures complete evidence-set recovery.
Current-only and missing-evidence queries are excluded from retrieval metrics and retained in the end-to-end evaluation.

\subsection{Memory-Writing Evaluation}

\paragraph{Transcription and temporal alignment.}
We report word error rate (WER) on the complete transcript and concatenated minimum-permutation WER (cpWER) for multi-speaker transcription~\citep{watanabe2020chime6challengetacklingmultispeakerspeech}.
For cpWER, utterances assigned to the same speaker are concatenated in temporal order, and the permutation between predicted and reference speaker streams that minimizes the total WER is used for scoring.
Temporal accuracy is measured with utterance-level intersection-over-union (IoU). Predicted and reference utterances are first matched using their textual similarity and temporal overlap, after which IoU is computed between the corresponding time intervals.
Thus, WER measures content transcription independently of speaker identity, whereas cpWER additionally penalizes incorrect assignment of recognized speech to speaker streams.

\paragraph{Speaker attribution.}
Speaker attribution is evaluated in an online setting using the timestamps and transcripts produced by the streaming ASR system.
For scoring only, predicted conversation-local speaker clusters are mapped one-to-one to reference speaker labels independently within each conversation.
We construct a cluster--speaker matrix from accumulated temporal overlap and use Hungarian assignment to maximize total matched overlap.
Reference labels are not used during online speaker assignment.
Time Acc is the fraction of jointly covered reference--prediction speech time assigned to the correct speaker,
\[
\mathrm{TimeAcc}
=
\frac{T_{\mathrm{correct}}}
     {T_{\mathrm{correct}}+T_{\mathrm{wrong}}}.
\]
Text Word Acc aligns predicted and reference utterances using text similarity and temporal overlap and computes speaker-assignment accuracy weighted by the number of aligned words.
Pair-F1 evaluates speaker clustering by considering whether pairs of utterances that belong to the same reference speaker are also assigned to the same predicted identity.
New-Spk F1 measures online detection of previously unseen speakers, while Known Acc measures assignment accuracy for utterances whose speaker has already been observed.
The operating threshold for assigning an utterance to an existing speaker is selected on the validation set and fixed before downstream evaluation.

\end{document}